\documentclass{article}
\usepackage{kr}

\usepackage{times}
\usepackage{soul}
\usepackage{url}
\usepackage[hidelinks]{hyperref}
\usepackage[utf8]{inputenc}
\usepackage[small]{caption}
\usepackage{graphicx}
\usepackage{amsmath}
\usepackage{amsthm}
\usepackage{booktabs}
\usepackage{todonotes}
\usepackage{amssymb}
\usepackage[ruled,vlined,linesnumbered]{algorithm2e}

\title{Knowledge- and Gradient-Guided Reinforcement Learning for Parametrized Action Markov Decision Processes}

\author{%
Jonas Ehrhardt$^{1,2}$\and
René Heesch$^{1,2}$\and
Oliver Niggemann$^{1,2}$ \\
\affiliations
$^1$HSU-AI Institute for Artificial Intelligence\\ 
$^2$Institute of Automation Technology\\
Helmut-Schmidt-University, Hamburg, Germany \\
\emails
\{firstname.lastname\}@hsu.hamburg
}

\begin{document}

\maketitle

\begin{abstract}
In this paper, we study Reinforcement Learning in Parametrized Action Markov Decision Processes (PAMDP), where each decision consists of a symbolic action and numerical parameters. 
In such settings Reinforcement Learning algorithms typically determine parameters with one-shot estimators, which makes their training sample inefficient.
Though in most  PAMDP environments explicit but incomplete knowledge (e.g., rules, safety constraints, or expert heuristics) is available, it is rarely directly used to increase the sample-efficiency of training Reinforcement Learning agents. 
We step into this gap and propose our novel Neuro-Symbolic Knowledge- and Gradient-Guided Reinforcement Learning (KGRL) algorithm.
KGRL uses domain knowledge in a Datalog knowledge base to derive the set of applicable actions and feasible parameters for a given state.
This allows it to prune non-applicable actions from the decision-space and constrain the parameter spaces of the remaining actions. We then use a gradient-based parameter refinement loop to estimate the optimal parameters during training and deployment of the agent. 
By recording activated rules along the trajectory, KGRL additionally provides local procedural explanations on the pruning of actions and constraining of parameters. 
Overall, KGRL guides the agent's exploration and deployment toward feasible and constraint-aware decisions, while increasing sample efficiency during training.
KGRL outperforms state-of-the-art RL baselines for PAMDPs in both, sample efficiency and episodic return.
\end{abstract}

\section{Introduction} \label{sec:introduction}
Reinforcement Learning (RL) is an alternative to symbolic decision-making algorithms for planning and control, when accurate symbolic models are unavailable or too costly to craft, e.g., due to continuous, stochastic, or high-dimensional dynamics~\cite{Sutton2018,Moerland2023}.
However, training RL agents requires an immense amount of experiences in the form of recorded training data or direct domain interactions~\cite{Yu2018,Duan2024}, while collecting such experience through exploration typically is costly or unsafe~\cite{Levine2020,DulacArnold2021}.
Yet, even in complex real-world domains, there is often explicit but incomplete domain knowledge (e.g., rules, safety constraints, or expert heuristics) available, which rules out unsafe or infeasible decisions without fully determining the optimal behavior~\cite{Achiam2017,Ghanadbashi2021}.
Hence, the central question we want to answer in this paper is: \textbf{Can explicit, yet incomplete, domain knowledge shape the training and deployment of an RL agent in order to (i) letting symbolic reasoning enforce known constraints, while (ii) leaving the learning to optimize what cannot be symbolically specified?}

We study this question in parametrized action spaces, where a decision involves not only choosing a symbolic action but also choosing numerical parameters that influence the action's effect~\cite{Masson2016}.
The hybrid structure is interesting, as it is common in many real-world decision-making problems, e.g., industrial control and planning~\cite{Zhu2021,Heesch2024}, where high-level actions are naturally symbolic (e.g., \texttt{open\_valve}, \texttt{start\_pump}, \texttt{fill\_tank}), but can accommodate numeric parameters (e.g., \texttt{valve\_opening}, \texttt{pump\_speed}, \texttt{fill\_height}).
The underlying problem of such scenarios is that the decision space is a union of disjoint continuous spaces, often of different dimensionality.
Approximations, like flattening all disjoint continuous spaces into a single continuous control space or discretizing all disjoint continuous spaces, can miss feasible or optimal parameter values or admit actions or parameters that violate constraints.
This calls for algorithms that are specifically tailored for such problems. 

Related work on RL in parametrized action spaces relies on one-shot estimators at decision time to estimate the parameters for actions~\cite{Masson2016,Hausknecht2016,Xiong2018,Bester2019}.
However, these approaches require numerous training episodes and do not directly address the risk of incurring out-of-distribution scenarios when deploying the RL agent.
Related work on integrating knowledge into RL agents typically focusses on crafting domain models~\cite{Silver2017,Silver2018,Moerland2023}, specifying goals in out-of-distribution scenarios~\cite{Ghanadbashi2021}, or constraining the agent for safe exploration and deployment \cite{Achiam2017,Alshiekh2018,Koenighofer2020,Koenighofer2025}.
Only very recently, few approaches started considering a Neuro-Symbolic perspective on RL that integrates symbolic decision-making as guardrail or augmentation of an RL policy \cite{Mazzi2023,Meli2024,Veronese2026}.
Yet, these approaches are typically only evaluated on discrete toy domains. 
To our knowledge, there is no work that integrates a structured, queryable knowledge base directly into the training and parameter estimation of an RL agent in a parametrized action space, to guide exploration and deployment toward feasible decisions and to improve training data efficiency.

To address this gap, we propose our novel, neuro-symbolic Knowledge- and Gradient-Guided Reinforcement Learning (KGRL) algorithm. 
KGRL introduces two coupled extensions to the DQN algorithm~\cite{Mnih2015}: 
A \textbf{knowledge base} which contains state-dependent rules that are evaluated before every action and parameter selection, and a \textbf{gradient-guided parameter refinement} loop that uses the gradient of the agent's action-value function to find the optimal parameters.
We ask the following research questions (RQs),

\noindent \textit{RQ1: Does integrating explicit domain knowledge from a symbolic knowledge base into the training of an RL agent increase sample efficiency and performance?}

Assuming that there is already a knowledge base of symbolic rules about the domain, we propose to use this knowledge in order to restrict the agent's exploration and deployment only to regions of the domain that are known to be not harmful.
Concretely, every step of the agent, we evaluate a Datalog knowledge base together with a symbolic state representation to derive the set of applicable actions and parameter constraints for the given state.
This allows us to streamline the exploration and deployment to only applicable actions and feasible parameters. 
\smallskip

\noindent \textit{RQ2: Can a gradient-guided parameter refinement increase the sample efficiency and performance of an RL agent, compared to one-shot estimators?}

In parametrized action spaces, the decision problem in each step consists of picking an action \textit{and} a continuous parameter. 
Existing algorithms typically use one-shot estimators, which predict in one forward pass a parameter value for a given state and action combination. 
However, in order to tune them into selecting optimal parameter picks and not violating domain constraints, their training becomes less sample efficient.
We hence propose a gradient-guided parameter refinement, which operates with the gradient of the action-value function and iteratively adapts the parameter values until they converge to an optimum. 
By using projected gradient ascent~\cite{Levitin1966}, we can integrate known parameter constraints from the knowledge base to exclude unrealistic parameter regions (e.g., temperatures below 0K) or parameters that violate known domain constraints.
\smallskip

\noindent \textit{RQ3: Can integrating explicit domain knowledge from a symbolic knowledge base into the deployment of an RL agent produce explanations of action and parameter choices?}

When evaluating a symbolic knowledge base and state representation on each step of an RL agent (cf. RQ1), KGRL can log the grounded rule instances and symbolic state facts that support the derivation of excluded actions and active parameter constraints from the decision space, as an explanation.
Such loggings pose local and procedural explanations~\cite{Krarup2019} along the trajectory of the agent.
This is particularly relevant for safety-critical, real-world applications and connects to work on explainable and safe RL~\cite{Alshiekh2018,Milani2024,Koenighofer2025}.

We evaluate our KGRL algorithm against baselines from literature~\cite{Masson2016,Hausknecht2016,Xiong2018,Li2022,Bester2019}.
KGRL outperforms all baselines, starting with higher returns and converging quicker to optimal results. 
Our main contributions are:
\begin{itemize}
    \item A formalization of knowledge-constrained Parametrized Action Markov Decision Processes that allows RL agents to include symbolic domain knowledge into their decision-making. 
    \item Our novel KGRL algorithm, that integrates reasoning over a symbolic knowledge base and a gradient-guided parameter refinement into its decision-making, leading to higher sample efficiency and performance. 
    \item An empirical evaluation of KGRL against state-of-the-art baselines.
    \item A demonstration on how local and procedural explanation traces can be logged from the stepwise evaluation of the knowledge base during the deployment of KGRL.
\end{itemize}
\section{Related Work} \label{sec:related-work}

\paragraph{RL in Parametrized Action Spaces}
was introduced in~\cite{Masson2016}, where the authors defined Parametrized Action Markov Decision Processes (PAMDPs) as the underlying formalism of the decision-making problems that incorporate numerical parameters. 
In a PAMDP the decision space is a union of disjoint continuous action subspaces, where each action subspace is defined by a symbolic action and a numerical parameter set~\cite{Masson2016}.
Early approaches treat this problem as two coupled decisions. 
In~\cite{Masson2016} a policy for selecting symbolic actions and a policy for selecting parameters are alternatingly updated during training. 
\cite{Fan2019} picked up on this idea and suggested an actor-critic approach, where separate policy heads for symbolic actions and numerical parameters operate on a joint state encoding in parallel.
Other approaches model the decision problem as a joint policy. 
For example,~\cite{Hausknecht2016} proposed an actor-critic approach, where one policy network outputs a symbolic action and an according parameter.
Contrary, ~\cite{Xiong2018} decouples the parameter estimation in a separate parameter estimator network that suggests parameter values which are then  jointly evaluated with the state by a Deep Q-Network~\cite{Mnih2015}. 
\cite{Bester2019} improved this approach by using batch-processing of individual parameter elements.
This mitigates confounding effects between parameters and yields a more robust training. 
Lastly,~\cite{Li2022} propose to leverage on a joint latent decision-space in which the decision-problem is solved, and from which symbolic actions and numerical parameters are decoded.
Across these approaches, action applicability and parameter feasibility are learnt implicitly, which yields sample-intensive exploration and can lead to infeasible parameters, if not prevented by the environment. 

\paragraph{Injecting Model- and Constraint-Knowledge into RL}
A fundamental way of exploiting and injecting knowledge about a domain into an RL algorithm lies in integrating it into a domain model or dynamics assumptions, as done in model-based RL~\cite{Moerland2023}.
For example,~\cite{Silver2017,Silver2018} demonstrated that a rule-based dynamics model of board games can help RL agents to surpass human performance in solving decision-making problems in such domains. 
Additionally, there is a line of work that incorporates constraints into RL, mainly for safe exploration and feasible exploitation. 
\cite{Achiam2017} proposed Constrained Policy Optimization (CPO), a policy-gradient approach that is regularized by an expected cost constraint and is designed to keep policy updates close to feasibility throughout the training, instead of only satisfying them at convergence. 
Also, ~\cite{Mazumder2022} regularize the exploration of RL agents by checking the permissibility of actions and constraining the action selection by this. 
However, these constraints are not derived from structured domain knowledge, but rely on learned functions before, or during training~\cite{Mazumder2022}.

\paragraph{Integrating Symbolic Reasoning into RL}
Most approaches that integrate symbolic reasoning for evaluating action applicability into the training of RL agents gathers in the domain of safe RL~\cite{Koenighofer2020,Koenighofer2025}.
\cite{Alshiekh2018} use temporal logic in a safety automaton to evaluate state-action pairs and acts as a shield to filter out unsafe actions. 
However, their algorithm only operates on discrete state- and action-spaces.
\cite{Fulton2018} use verified model constraints of a hybrid system model to create a monitor for maintaining safe system behavior.
\cite{Anderson2020} use projection into a predefined verifiable symbolic policy class, to allow for safe exploration. 
Inarguably the closest to our work is \cite{Veronese2026} and \cite{Han2026}. 
\cite{Veronese2026} propose to augment a DQN agent with a soft symbolic policy that is obtained from evaluating Answer Set Programming rules with a symbolic state abstraction, while \cite{Han2026} use Probabilistic Sentential Decision Diagrams to mask unsuitable actions. 
However, their approaches remain bound to discrete domains \cite{Veronese2026,Han2026}. 

While overall there is work that combines symbolic knowledge with RL, to our knowledge there is no work focussing on the combination of incomplete symbolic knowledge and RL in parametrized action spaces.

\section{Problem Formalization} \label{sec:problem}

In this section, we formalize how RL in parametrized action spaces can be connected to symbolic reasoning. 
Therefore, we build on Parametrized Action Markov Decision Processes (PAMDPs)~\cite{Masson2016} and enrich them with a Datalog knowledge base and an abstraction function. 
This allows us to abstract numeric PAMDP states into symbolic representation and evaluate them together with the knowledge base to derive applicable actions and parameter constraints. 

\subsection{Parametrized Action Markov Decision Processes}
A PAMDP is a tuple
\begin{equation}
    \langle S, A, \{\Psi_a\}_{a \in A}, T, r, \gamma \rangle
\end{equation}
where $S \subseteq \mathbb{R}^n$ is the continuous state space, 
$A$ is a finite set of symbolic actions. 
Each action $a \in A$ is associated with a continuous parameter space $\Psi_a \subseteq \mathbb{R}^{m_a}$.
Hence, the hybrid parametrized action space is the set
\begin{equation}
    \mathcal{A} = \{ (a, \psi) | a \in A, \psi \in \Psi_a \}.
\end{equation}
$T$ is the transition function $T = P(s'|s, a, \psi)$ that describes the probability of transitioning into state $s' \in S$ given state $s \in S$, action $a \in A$ and a parameter $\psi \in \Psi_a$.
$r$ is the reward function $r: S \mathcal{ \times A \times S \rightarrow \mathbb{R}}$ that returns the scalar reward when transitioning from $s$ to $s'$.
$\gamma \in [0,1)$ is a discount factor.

A policy is a function $\pi: S \rightarrow \mathcal{A}$.
Executing $\pi$ leads to a trajectory $(s, (a, \psi), r, d, s', (a', \psi'), r', d' ...)$, where $\pi(s) = (a, \psi)$, $s' \sim T(\cdot | s, a, \psi)$, $r = r(s, a, \psi)$, and $d \in \{0,1\}$ indicates whether an episode is done, $d=1$, or not $d=0$.
The performance of a policy is measured by its expected discounted return
\begin{equation}
    J(\pi) = \mathbb{E}_\pi \left[ \sum_{t=0}^\infty \gamma^t r(s_t, a_t, \psi_t) \right].
\end{equation}

\subsection{Symbolic Abstraction and Knowledge Base}
In real-world domains, there is often explicit, yet incomplete domain knowledge in the form of rules, safety constraints, and expert heuristics available. 
This knowledge typically excludes unsafe or infeasible decisions, but does not specify optimal behavior.
One way of representing such domain knowledge is in the form of a Datalog knowledge base $\mathcal{K}$.

\paragraph{Symbolic Abstraction}
To connect the numeric PAMDP states $s \in S$ with the symbolic knowledge base $\mathcal{K}$, we need an abstraction from numeric to symbolic state representations. 
Therefore, we assume a fixed finite set $\mathcal{F}$ of propositional atoms that allow for describing the features of $S$. 
We assume that there is a given abstraction function $\alpha$ which maps numeric PAMDP states into a symbolic representation. 
\begin{equation}
    \alpha : S \rightarrow 2^\mathcal{F}.
\end{equation}
For example, $\alpha$ can be defined by numeric threshold tests over $s$, or learned~\cite{Ciatto2024}.
This allows us to obtain the symbolic state representation $\alpha(s)$ as the set of propositional feature atoms that hold true in $s$.

For example, consider $s$ to contain a feature tank level $\texttt{level}(s) \in [0,42]$.
$\alpha$ may be defined by threshold tests, like $\texttt{tank\_near\_full} \in \alpha(s) \iff \texttt{level}(s) \geq 41$ or $\texttt{tank\_near\_empty} \in \alpha(s) \iff \texttt{level}(s) \leq 1$, where $\texttt{tank\_near\_full}, \texttt{tank\_near\_empty} \in \mathcal{F}$. 
Given a state $s$, we could obtain the symbolic state representation by applying $\alpha$ to $s$. E.g., $\texttt{level}(s) = 41.25$ would yield $\alpha(s) = \{ \texttt{tank\_near\_full} \}$, while $\texttt{level}(s) = 0.25$ yields $\alpha(s) = \{ \texttt{tank\_near\_empty} \}$.

\paragraph{Knowledge Base and Evaluation}
We assume that $\mathcal{K}$ is a Datalog knowledge base and consists of a finite set of rules with negation-as-failure in the form of 
\begin{equation}
    h \gets b_1,\ldots,b_x,\ \texttt{not }b_{x+1},\ldots,\ \texttt{not }b_z,
\end{equation}
where $h$ and all $b_i$ are atoms over a fixed predicate vocabulary, and \texttt{not} denotes negation-as-failure. 

In addition to the rules in $\mathcal{K}$ and the symbolic state representation from $\alpha(s)$, we work with static declarations of actions and constraints.
When $A$ is the finite set of symbolic actions from the PAMDP and $C$ is the finite set of parameter constraint identifiers, we can denote the static declarations 
\begin{align*}
    Act &:=\{\texttt{action}(a) | a \in A\} \\
    Con &:=\{\texttt{constr}(c) | c \in C \}.
\end{align*}
As $\mathcal{K}$ is stratified, $\mathcal{K} \ \cup \ \alpha(s) \ \cup \ Act \ \cup Con$ has a unique intended model for every state $s$. 
We write 
\begin{equation}
    \mathcal{K} \ \cup \ \alpha(s) \ \cup \ Act \ \cup Con \models q, 
\end{equation}
to denote that the atom $q$ is entailed in state $s$.

\paragraph{Applicable Actions}
Evaluating the program above allows us to obtain the set of actions that are applicable in state $s$. 
Therefore, we use the predicates $\texttt{app}(a)$ to mark applicability of actions and $\texttt{napp}(a)$ to mark non-applicability of actions.
The set of applicable actions can be defined as
\begin{equation}
    \bar{A}(s) = \{a \in A | \mathcal{K} \cup \alpha(s)\cup Act\cup Con \models \texttt{app}(a) \}.
\end{equation}
As we assume that the encoded knowledge in $\mathcal{K}$ is incomplete, we adopt a default applicability scheme\footnote{This resembles reasoning with closed-world assumption. Yet, as we restrict this only to the applicability of actions and activeness constraints (and not on predicates and unknown facts), we consider it a default policy rather than a strict closed-world assumption.}. 
An action $a$ is considered applicable unless the predicate $\texttt{napp}(a)$ (non-applicable) can be derived.
We define the default action applicability by the rule 
\begin{equation}
    \texttt{app}(a) \leftarrow \texttt{action}(a), \texttt{not napp}(a).
\end{equation}

\paragraph{Feasible Parameters}
Evaluating the program additionally determines which parameter constraints are active for an action $a$ in state $s$. 
We use the predicate $\texttt{active}(a,c)$ to denote that the constraint identifier $c \in C$ is active for $a \in A$, and define
\begin{align}
    C_{\mathcal{K}}(s,a) = \{ c \in C \mid & \mathcal{K} \cup \alpha(s)\cup Act\cup Con \\ &\models \texttt{active}(a,c)\}. \nonumber
\end{align}
Each constraint identifier $c$ is grounded by a numeric feasibility condition for $a$ in $s$, inducing the feasible subset
\begin{equation}
    \Psi_{a,c}(s) \subseteq \Psi_a.
\end{equation}
The feasible parameter region for $a$ in $s$ is then
\begin{align}
    \bar{\Psi}_a(s) = & \bigcap_{c\in C_{\mathcal{K}}(s,a)} \Psi_{a,c}(s),\\
    &\text{where } \bigcap_{c\in\emptyset} \Psi_{a,c}(s) = \Psi_a. \nonumber
\end{align}

Together, $\bar{A}(s)$ and $\{\bar{\Psi}_a(s)\}_{a \in A}$ characterize the decision space that remains feasible for the RL agent after incorporating the symbolic domain knowledge.

\subsection{Knowledge-Constrained PAMDP}
As we can obtain the set of feasible parametrized actions in a state $s$ as
\begin{align}
    \bar{\mathcal{A}}_{\mathcal{K}}(s) = \{ (a,\psi) | & a \in \bar{A}(s), \psi \in \bar{\Psi}_a(s) \}, \nonumber
\end{align}
we can now construct a knowledge-constrained PAMDP as a tuple
\begin{equation}
    \langle  S, A, \{\Psi_a\}_{a \in A}, T, r, \gamma, \alpha, \mathcal{K}  \rangle.
\end{equation}

A policy $\pi$ is $\mathcal{K}$-feasible if it only selects decisions that are admissible by the knowledge base
\begin{equation}
    \forall s \in S: \pi(s) \in \bar{\mathcal{A}}_{\mathcal{K}}(s).
\end{equation}

When $\Pi_{\mathcal{K}}$ describes the set of all $\mathcal{K}$-feasible policies, then a knowledge-constrained RL problem induced by a knowledge-constrained PAMDP lies in finding a policy $\pi^\star \in \Pi_{\mathcal{K}}$ that maximizes the return
\begin{equation}
    \pi^\star \in \arg \max_{\pi \in \Pi_{\mathcal{K}}} J(\pi).
\end{equation}

\section{Solution} \label{sec:solution}

\begin{figure}
    \centering
    \includegraphics[width=0.95\linewidth]{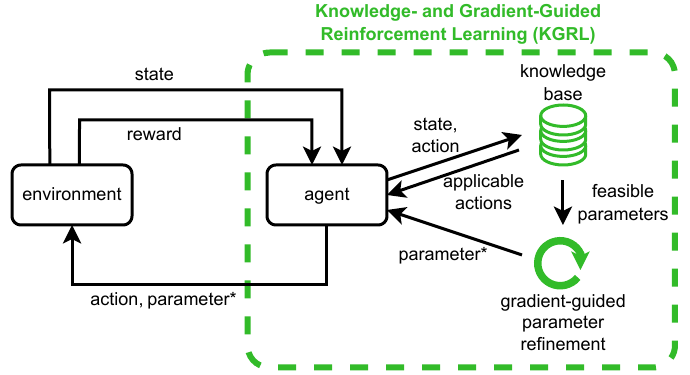}
    \caption{We propose the novel Neuro-Symbolic Knowledge- and Gradient-Guided Reinforcement Learning (KGRL) algorithm. 
    KGRL uses a knowledge base $\mathcal{K}$ at each decision to exclude non-applicable actions and derive the feasible parameter space for each applicable action.
    Using gradient-based parameter optimization, KGRL can refine an initial suboptimal parameter guess at decision time into an optimal parameter estimate.}
    \label{fig:teaser}
\end{figure}

In this section, we present our novel, Neuro-Symbolic Knowledge- and Gradient-Guided Reinforcement Learning (KGRL) algorithm (cf. Figure \ref{fig:teaser}); our solution to a knowledge-constrained PAMDP.
KGRL is based on the Deep Q-Network (DQN) algorithm by~\cite{Mnih2015}, which approximates the action-value function of a domain in a deep Neural Network $Q_{\theta}$.
The key idea of KGRL is to use symbolic domain knowledge from a Datalog knowledge base to prune non-applicable actions and restrict parameters to feasible regions. 
Therefore, we extend DQN with two additional mechanisms: 
(i) A Datalog knowledge base $\mathcal{K}$ that can derive the set of applicable actions and feasible parameters regions for every state,
(ii) $\textsc{ParamOpt}$, a gradient-guided parameter refinement loop that uses the gradient of $Q_{\theta}$ to estimate optimal parameters within the feasible parameter regions.

In each step, KGRL evaluates $\mathcal{K}$ and the symbolic representation of the current state $\alpha(s)$ to derive the set of applicable actions $\bar{A}(s)$ and feasible parameters $\{\Psi_a\}_{a \in \bar{A}(s)}(s)$.
This allows KGRL to prune non-applicable actions and infeasible parameters from the decision space.
After pruning the action space, $\textsc{ParamOpt}$ uses the gradient information of $Q_\theta$ to perform a gradient-guided parameter refinement as an optimization over the feasible parameter spaces of the available actions. 
At the end of each step, KGRL greedily selects the action-parameter tuple with the highest action-value.

\subsection{Knowledge- and Gradient-Guided Q-Learning}
Algorithm~\ref{alg:kgrl} summarizes the training loop of our KGRL algorithm.
To stabilize training, we use a double DQN setup, where two DQNs, an online DQN and a target DQN, are synchronized periodically \cite{Mnih2015}.
KGRL starts with initializing the weights of the online and target DQN $\theta$ and $\theta^-$, as well as the replay buffer $\mathcal{D}$, and the environment $\mathcal{E}$ (Alg.~\ref{alg:kgrl}, lines 1-3). 
Until the maximum budget of training steps $T_{\max}$ is reached, it derives the set of applicable actions $\bar{A}(s)$ and feasible parameters $\{\Psi_a\}_{a \in \bar{A}(s)}(s)$ for the current state $s$ (Alg.~\ref{alg:kgrl}, line 5). 
Based on $\bar{A}(s)$ and $\{\Psi_a\}_{a \in \bar{A}(s)}(s)$ KGRL either performs exploration (Alg.~\ref{alg:kgrl}, line 6-8) or exploits the existing policy (Alg.~\ref{alg:kgrl}, line 9-12), using the $\textsc{ParamOpt}$ algorithm (cf. Alg.~\ref{alg:paramopt}).
The DQNs are updated with mini-batches $B$ that are sampled from $\mathcal{D}$ (Alg.~\ref{alg:kgrl}, line 15).
The targets are computed as
\begin{align} \label{eq:target}
    &y = r + \gamma(1 - d') \underset{a' \in \bar{A}(s')}{\arg\max} \max_{\psi' \in \bar{\Psi}_a(s')} Q_{\theta^-}(s', a', \psi').
\end{align} (Alg.~\ref{alg:kgrl}, lines 18-21).
The weights of the online DQN are updated via gradient descent on the squared Bellman Error (Alg.~\ref{alg:kgrl}, lines 22-23). 
Periodically, the weights of the target DQN are synchronized with the weights of the online DQN (Alg.~\ref{alg:kgrl}, line 25).

\begin{algorithm}[h!]
\caption{KGRL}\label{alg:kgrl}
\DontPrintSemicolon
\SetKwInOut{Require}{Require}
\Require{
\begin{tabular}[t]{@{}l@{\quad}l@{}}
$\mathcal{E}$ & \tcp{environment}\\
$\mathcal{D}$ & \tcp{replay buffer} \\
$Q_\theta, Q_{\theta^-}$ & \tcp{online \& target DQN}\\
$\mathcal{K}$ & \tcp{knowledge base}\\
$\alpha$ & \tcp{abstraction function} \\
$\epsilon$ & \tcp{exploration rate} \\
$\beta$ & \tcp{learning rate} \\
$\gamma$ & \tcp{discount factor}\\
\end{tabular}
}

$\theta, \theta^- \gets \text{init}()$\;
$\mathcal{D} \gets \mathcal{D}.\text{init}()$\;
$s \gets \mathcal{E}.\text{init}()$\;
\For{$t \gets 1$ \KwTo $T_{\max}$}{
    $\bar{A}(s), \{\bar{\Psi}_a\}_{a \in \bar{A}(s)}(s) \gets \textsc{evaluate}(\mathcal{K}, \alpha(s))$\;

    \tcp{exploration}
    \If{$\text{random}() < \epsilon$}{
        $a \gets \bar{A}(s).\text{sample}()$\;
        $\psi \gets \bar{\Psi}_{a}(s).\text{sample}()$\;
    }
    
    \tcp{exploitation}
    \Else{
        \ForEach{$a \in \bar{A}(s)$}{
            $\tilde{\psi}_{a}^\star \leftarrow \textsc{ParamOpt}(Q_\theta,s,a,\bar{\Psi}_a(s))$\;
        }
        $(a,\tilde{\psi}_{a}^\star)\leftarrow \underset{a \in \bar{A}(s)}{\arg \max} Q_\theta(s,a,\tilde{\psi}_{a}^\star)$\;

    }
    
    $r, s', d' \gets \mathcal{E}(s, a, \tilde{\psi}_{a}^\star)$
    $\mathcal{D} \gets \mathcal{D}.\text{append}((s, a, \tilde{\psi}_{a}^\star, r, d', s'))$ \;

    \tcp{update DQNs}
    \If{$\text{update condition holds}$}{
        \tcp{sample mini-batch}
        $B \gets \mathcal{D}.\text{sample}()$ \; 
        \ForEach{$(s_i, a_i, \psi_i, r_i, d'_i, s'_{i}) \in B$}{
            $\bar{A}(s_i'), \{\bar{\Psi}_{a'}\}_{a' \in \bar{A}(s_i')}(s'_i) \gets \textsc{evaluate}(\mathcal{K}, \alpha(s))$\;
            \ForEach{$a' \in \bar{A}(s_{i}')$}{
                $\tilde{\psi}^{'\star}_{i,a'} \gets\textsc{ParamOpt}(Q_\theta, s_{i}', a', \bar{\Psi}_{a'}(s_{i}'))$
            }
            $a_i' \gets \underset{a' \in \bar{A}(s_i')}{\arg \max} Q_{\theta}(s_{i}', a',\tilde{\psi}_{i,a'}^{'\star}) $ \;
            
            $y_i \gets r_i + \gamma (1-d_i') Q_{\theta^-}(s_i', a_i', \tilde{\psi}_{i,a_i'}^{'\star})$ \; 
        }
    $\mathcal{L} \gets \frac{1}{|B|} \sum_i (y_i - Q_{\theta}(s_i, a_i, \psi_i))^2$ \;
    $\theta \gets \theta - \beta \nabla_{\theta}\mathcal{L}$ \;
    
    \If{$\text{sync condition}$}{
    \tcp{sync online and target DQN}
    $\theta^- \gets \theta$\; 
    }
    }
    }
\end{algorithm}

\subsection{Gradient-Guided Parameter Refinement}
As we assume a numerical parameter space $\bar{\Psi}_{a}(s)$, calculating the optimal parameter $\psi^\star \in \bar{\Psi}_{a}(s)$ over the highly non-convex action-value function $Q_{\theta}$ is intractable. 
Because search can be computationally expensive, we propose to leverage on the differentiability of $Q_\theta$ for estimating $\psi^\star$~\cite{Ehrhardt2025}. 
Precisely, we suggest our $\textsc{ParamOpt}$ algorithm, which performs projected gradient ascent~\cite{Levitin1966} along the gradient of $Q_\theta$ with respect to $\psi$ (cf. Algorithm \ref{alg:paramopt}).
This allows us to refine an initial parameter guess $\hat{\psi}$ until it converges to an optimal parameter value. 
As we are using a projection function $\text{Proj}_{\bar{\Psi}_a(s)}(\cdot)$ we can remain within the bounds of the feasible parameter space. 
However, due to the non-convexity of $Q_{\theta}$, we cannot guarantee to reach $\psi^\star$.
Hence, we denote the result of $\textsc{ParamOpt}$ as $\tilde{\psi}^\star$.
$\textsc{ParamOpt}$ starts with an initial parameter guess $\hat{\psi}$, which can either be zeros, random values, or expert heuristics (Alg.~\ref{alg:paramopt}, line 1). 
It then calculates the gradient of the DQN with respect to $\psi$ (Alg.~\ref{alg:paramopt}, line 3) and performs projected gradient ascent until the gradient norm falls below a stopping threshold $\xi$ (Alg.~\ref{alg:paramopt}, line 5) or until the maximum number of optimization steps $U_{\max}$ (Alg.~\ref{alg:paramopt}, line 2) is reached. 

\begin{algorithm}[h!]

\caption{\textsc{ParamOpt}} \label{alg:paramopt}
\DontPrintSemicolon
\SetKwInOut{Require}{Require}
\Require{
\begin{tabular}[t]{@{}l@{\quad}l@{}}
$s$                        & \tcp{state}\\
$a \in \bar{A}(s)$         & \tcp{action}\\
$\bar{\Psi}_{a}(s)$        & \tcp{feasible parameters}\\
$Q_{\theta}$               & \tcp{DQN}\\
$\zeta$                    & \tcp{learning rate}\\
$\xi$                      & \tcp{stopping threshold}
\end{tabular}
}
    
    $\hat{\psi} \leftarrow \text{init}()$ \tcp{initial parameter guess}
    
    \For{$u \leftarrow 1$ \KwTo $U_{\max}$}{
        \tcp{backprop wrt. parameters}
        $g_{\hat{\psi}} \leftarrow \nabla_{\hat{\psi}} Q_\theta(s, a, \hat{\psi})$ \; 
        \tcp{projected gradient ascent}
        $\hat{\psi} \leftarrow \text{Proj}_{\bar{\Psi}_{a}(s)}(\hat{\psi} + \zeta \, g_{\hat{\psi}})$ \; 
        \If{$||g_{\hat{\psi}}|| < \xi$}{
            $\textbf{break}$ \;
        }
        }
    \Return $\tilde{\psi}^* \leftarrow \hat{\psi}$

\end{algorithm}

\subsection{Explanation Traces}
Since KGRL evaluates $\mathcal{K}$ together with $\alpha(s)$ at every decision step, it can record the supporting derivations as an explanation trace along the trajectory. 
At each time step, the trace contains the symbolic facts from $\alpha(s)$ and the rules in $\mathcal{K}$ that derive $\texttt{app}(a)$, $\texttt{napp}(a)$, and $\texttt{active}(a,c)$. 
This makes explicit why actions were pruned or admitted and why the parameter search was restricted to a particular feasible region. 
The resulting trace constitutes a local procedural explanation~\cite{Krarup2019}, as it explains parts of the concrete decision process at each state rather than the policy as a whole.
\section{Evaluation} \label{sec:evaluation}
In this section, we present the empirical evaluation of our KGRL algorithm. 
We aim to answer our research questions with two experiments. 
First, we compare KGRL and its ablations against baseline RL algorithms for PAMDPs.
Second, we evaluate KGRL's knowledge base utilization during application in order to derive and evaluate explanation traces.

\subsection{Experimental Setup}
We evaluate our KGRL algorithm against five baseline RL algorithms for PAMDPs from literature: 
QPAMDP~\cite{Masson2016}, 
PA-DDPG~\cite{Hausknecht2016}, 
P-DQN~\cite{Xiong2018},
MP-DQN~\cite{Bester2019},
and HyAR~\cite{Li2022}.

\begin{figure}[h!]
    \centering
    \includegraphics[width=0.8\linewidth]{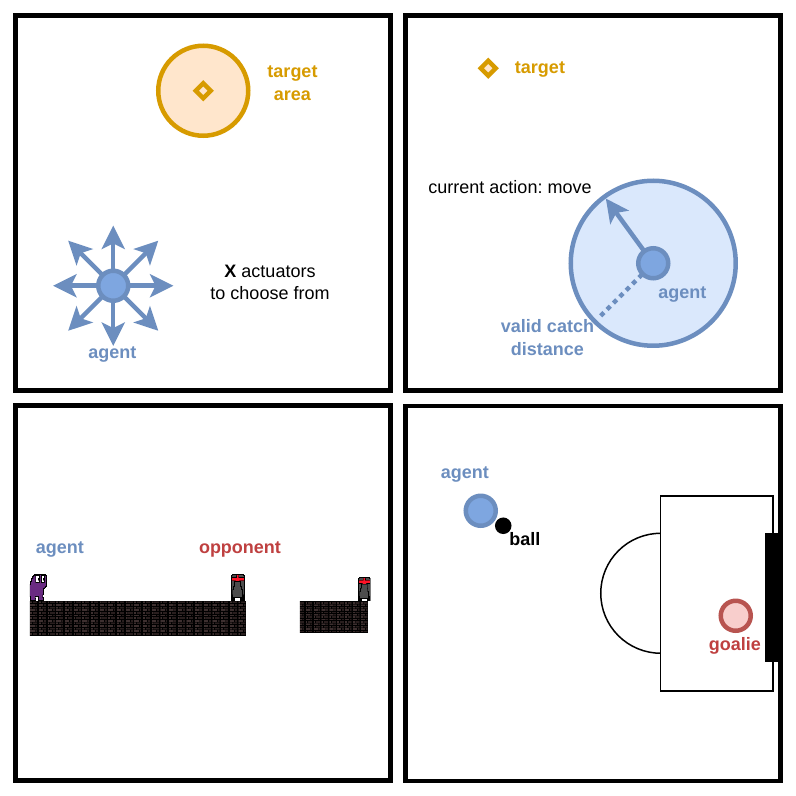}
    \caption{The Figure shows our evaluation PAMDP domains. 
    We evaluated our algorithm on the \textit{HardMoveX} (upper left), \textit{CatchPoint} (upper right), \textit{Platform} (lower left), and \textit{\textit{HardGoal}} (lower right) domains.}
    \label{fig:domains}
\end{figure}

As evaluation domains we use the \textit{HardMoveX}, \textit{CatchPoint}, \textit{\textit{HardGoal}} domains from~\cite{Li2022}, as well as the \textit{Platform} domain from~\cite{Masson2016} (cf. Figure \ref{fig:domains}).
All domains have parametrized action spaces that consist of a finite set of symbolic actions and associated numerical parameters.
For each domain we crafted a set of simple rules to utilize in KGRL.

\textit{HardMoveX} describes a 2D navigational domain in which the agent must reach a target. 
The agent has $n=2X$ symbolic actions $\texttt{moveX}(\psi)$ it can choose from.
Each symbolic action is pointing in one specific direction and is parametrizable by a magnitude of $\psi \in [0,1]$.
Rules in $\mathcal{K}$ prune actions that point away from the goal in a larger angle than 90°, and cap parameters based on the distance to the goal.
\textit{HardMoveX} is especially suitable to evaluate the capability to handle large action spaces. 
We tested our algorithm on \textit{HardMove4}, \textit{HardMove8}, and \textit{HardMove10}.

The \textit{CatchPoint} domain is a 2D reach-and-catch task. 
The agent can $\texttt{move}(\psi), \psi \in [-1,1]$ and $\texttt{catch}(\psi), \psi \in [0,1]$, both parameters are mapped to an angle. 
Rules in $\mathcal{K}$ prune the moving action when the agent is already in target reach, and prune the catch action when catching is impossible. 

The \textit{\textit{HardGoal}} domain is a simplified soccer benchmark, where the agent aims to shoot a goal against a goalkeeper. 
Therefore, it can choose from ten shooting actions $\texttt{a0}(\psi)$ to $\texttt{a9}(\psi), \psi \in [-1, 1]$. 
The goalkeeper is continuously moving. 
Rules in $\mathcal{K}$ prune the lower half actions, when the goalkeeper is in the left goal corner and upper-half actions, when the goalkeeper is in the right corner, as well as any action that is blocked by the keeper.

\begin{table*}[h!]
\centering
\setlength{\tabcolsep}{4pt}
\caption{
The table compares the AULC (the higher, the better) of our KGRL algorithm and its ablations against the baseline algorithms on different evaluation domains. 
We compare ablations of KGRL using no gradient-based parameter refinement (nogg), no knowledge-guidance (nokb), or neither (nogg nokb) during training (train) or deployment (eval). 
As evaluation metric, we report average episodic return on 16 evaluation problems $\pm$ one standard deviation across eight different seeds.}\label{tab:results:exp1}
\begin{tabular}{lllllll}
\toprule
Algorithm                   & HardMove4         & HardMove8         & HardMove10        & CatchPoint       & HardGoal         & Platform          \\
\midrule
QPAMDP                      & -13.63 $\pm$ 0.30 & -13.49 $\pm$ 0.14 & -13.56 $\pm$ 0.23 & -9.67 $\pm$ 0.29 & 39.20 $\pm$ 2.34 & 0.03 $\pm$ 0.05 \\
PA-DDPG                     & -14.12 $\pm$ 2.49 & -14.63 $\pm$ 3.14 & -16.09 $\pm$ 6.98 & -7.40 $\pm$ 2.48 & 32.86 $\pm$ 8.86 & 0.09 $\pm$ 0.09 \\
P-DQN                       & -12.84 $\pm$ 1.24 & -10.37 $\pm$ 2.79 & -9.74 $\pm$ 2.95  & -9.79 $\pm$ 0.34 & 48.17 $\pm$ 0.48 & 0.13 $\pm$ 0.07 \\
HyAR                        & -7.49 $\pm$ 5.66  & -5.15 $\pm$ 0.40  & -5.08 $\pm$ 0.40  & -8.19 $\pm$ 1.77 & 42.04 $\pm$ 5.30 & 0.08 $\pm$ 0.04 \\
MP-DQN                      & -3.48 $\pm$ 0.48  & -3.47 $\pm$ 0.30  & -3.28 $\pm$ 0.36  & -5.23 $\pm$ 0.31 & 47.62 $\pm$ 0.64 & 0.19 $\pm$ 0.06 \\
KGRL                        & \textbf{-1.64 $\pm$ 0.21}  & \textbf{-1.44 $\pm$ 0.13}  & \textbf{-1.36 $\pm$ 0.06}  & \textbf{-1.70 $\pm$ 0.00} & \textbf{49.76 $\pm$ 0.07} & \textbf{0.22 $\pm$ 0.02} \\
\midrule
KGRL (train nogg)           & \textbf{-1.43 $\pm$ 0.13}  & \textbf{-1.34 $\pm$ 0.10}  & \textbf{-1.49 $\pm$ 0.18}  & -1.69 $\pm$ 0.00 & 49.74 $\pm$ 0.08 & 0.21 $\pm$ 0.02 \\
KGRL (train nokb)           & -1.84 $\pm$ 0.17  & -1.56 $\pm$ 0.09  & -1.62 $\pm$ 0.14  & -1.75 $\pm$ 0.02 & \textbf{49.87 $\pm$ 0.06} & 0.21 $\pm$ 0.02 \\
KGRL (train nogg nokb)      & -1.68 $\pm$ 0.19  & -1.81 $\pm$ 0.21  & -1.74 $\pm$ 0.14  & -1.74 $\pm$ 0.02 & 49.86 $\pm$ 0.05 & \textbf{0.23 $\pm$ 0.02} \\
KGRL (eval nogg)            & -1.68 $\pm$ 0.17  & -1.53 $\pm$ 0.15  & -1.50 $\pm$ 0.13  & \textbf{-1.66 $\pm$ 0.00} & 49.72 $\pm$ 0.06 & 0.21 $\pm$ 0.03 \\
KGRL (eval nokb)            & -4.33 $\pm$ 0.43  & -4.17 $\pm$ 0.35  & -4.41 $\pm$ 0.55  & -6.28 $\pm$ 0.42 & 34.00 $\pm$ 1.22 & 0.19 $\pm$ 0.02 \\
KGRL (eval nogg nokb)       & -4.81 $\pm$ 0.36  & -4.99 $\pm$ 0.46  & -5.02 $\pm$ 0.28  & -6.34 $\pm$ 0.44 & 34.68 $\pm$ 1.85 & 0.20 $\pm$ 0.03 \\
KGRL (train eval nogg)      & -1.74 $\pm$ 0.22  & -1.53 $\pm$ 0.12  & -1.60 $\pm$ 0.15  & \textbf{-1.66 $\pm$ 0.00} & 49.66 $\pm$ 0.05 & 0.22 $\pm$ 0.03 \\
KGRL (train eval nokb) & -4.19 $\pm$ 0.40  & -4.50 $\pm$ 0.31  & -4.64 $\pm$ 0.44  & -5.49 $\pm$ 0.35 & 43.23 $\pm$ 0.86 & 0.22 $\pm$ 0.03 \\
KGRL (train eval nogg nokb) & -5.21 $\pm$ 0.43  & -5.33 $\pm$ 0.41  & -4.69 $\pm$ 0.29  & -6.22 $\pm$ 0.40 & 42.55 $\pm$ 1.50 & \textbf{0.23 $\pm$ 0.02} \\
\bottomrule
\end{tabular}
\end{table*}

\begin{figure*}[h!]
\centering
\includegraphics[width=0.24\linewidth]{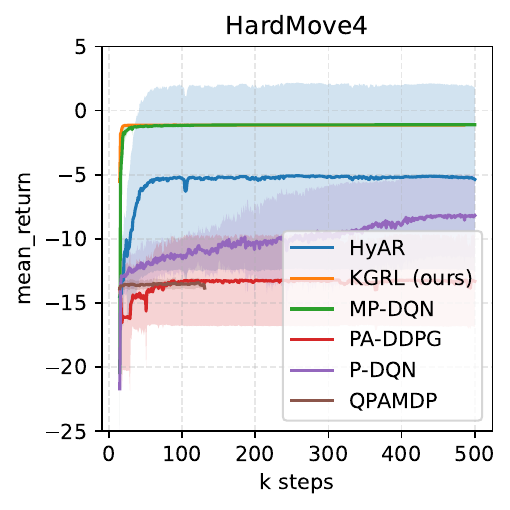}
\includegraphics[width=0.24\linewidth]{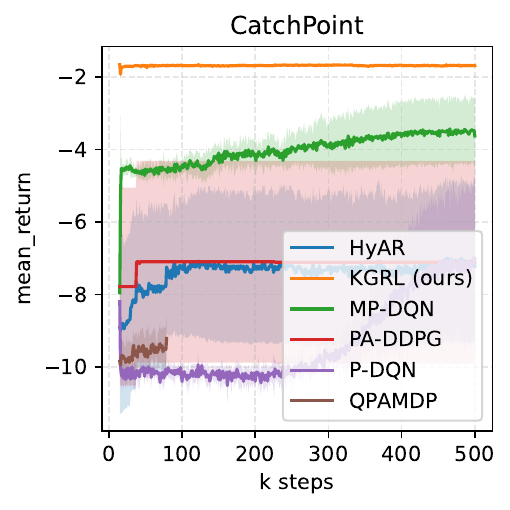} 
\includegraphics[width=0.24\linewidth]{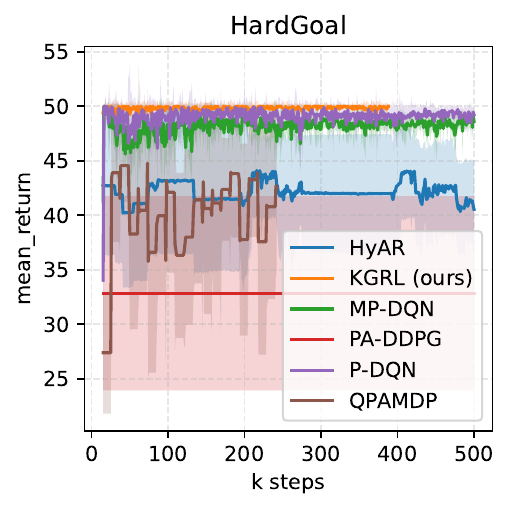}
\includegraphics[width=0.24\linewidth]{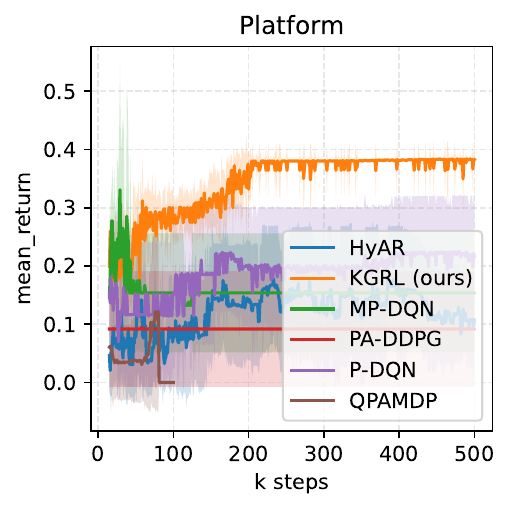}
\caption{The Figure shows the learning curves of KGRL and the baselines on the \textit{HardMove4, CatchPoint, HardGoal}, and \textit{Platform} evaluation domains. The x-axis shows training progress in environment steps, and the y-axis shows the mean episodic return $\pm$ one standard deviation across eight different seeds on the evaluation tasks (shaded area). Steeper slopes indicate faster learning and higher values indicate better learned policies.} \label{fig:results:exp1}
\end{figure*}

The \textit{Platform} domain describes a platform traversal task with gaps and opponents.
The agent can $\texttt{run}(\psi), \psi \in [0,30]$, $\texttt{hop}(\psi), \psi \in [0,720]$, and $\texttt{leap}(\psi), \psi \in [0,430]$. 
Rules in $\mathcal{K}$ prune actions that endanger the agent by moving too close to a gap or the opponent, as well as constrain parameters, based on features like enemy closeness, high velocity, or near gaps. 

For creating our experimental setup, we followed the experimental design guidelines for empirical Machine Learning research by~\cite{Vranjes2024}.
Therefore, we compared the performance of our KGRL algorithm against the baseline algorithms on the evaluation domains. 
We trained all algorithms with a fixed budget of 500,000 training steps and a maximum computation wall time of 12 hours. 
To rule out lucky initialization, we repeated the trainings on eight different seeds. 
Every 1,000 steps, we performed an evaluation run on 16 differently instantiated evaluation problems.
\bigskip

The code for reproducing the results, including the specific rules sets, can be found under \url{https://github.com/j-ehrhardt/kgrl}.

\subsection{KGRL vs. Baselines}
In the first experiment, we evaluated the Area Under the Learning Curve (AULC) of our KGRL algortihm against the baseline algorithms. 
We chose the AULC as it combines algorithmic performance and convergence speed in a single metric.
The results are reported in the upper part of Table~\ref{tab:results:exp1} and Figure~\ref{fig:results:exp1}.
We could show that across all evaluation domains, KGRL consistently achieves the highest average AULC across the eight different seeds. 
Additionally, we could show that KGRL already starts with higher episodic returns and converges quicker to higher return values, with lower variance across the seeds than the baseline algorithms (cf. Figure \ref{fig:results:exp1}).
Both findings support our RQ1 that integrating explicit but incomplete knowledge into the training and application of an RL agent improves learning efficiency and performance.


Additionally, we evaluated ablations of KGRL, to estimate the influence of the knowledge and gradient guidance.
Therefore, we removed the knowledge base (nokb) or replaced $\textsc{ParamOpt}$ algorithm with a mean estimate of the parameter range (nogg).
We ablated the knowledge base or $\textsc{ParamOpt}$ only during training (train), during evaluation (eval), or during both (train eval). 
The results are reported in the lower part of Table \ref{tab:results:exp1}.
Ablating knowledge- and gradient-guidance showed mixed influence on the AULC and a strong dependence on the evaluation. 
While ablating the knowledge base during evaluation showed a consistently worse AULC, ablating the gradient-guidance during training led to better results in the \textit{HardMoveX} domains and comparable results to the full version of KGRL in the other domains. 
Also in the \textit{CatchPoint} domain ablating the gradient guidance during evaluation or training and evaluation lead to better AULC than the full version of KGRL. 
However, removing knowledge- and gradient-guidance during training and evaluation from KGRL, rendered consistently worse results than keeping them, except for the \textit{Platform} domain, which generally showed little variance across the ablations.

\subsection{Knowledge Base Utilization and Explanation Traces}
In addition to the algorithm performance, we also quantified how strongly KGRL utilizes the knowledge base during application, and whether its use is substantial enough to support rule-grounded explanations. 
Therefore, we evaluated the knowledge base utilization $\mathcal{K}$-utilization, and the mean action pruning rate $A$-pruning as metrics. 
The knowledge base utilization describes evaluations of the knowledge base that resulted either in activations of pruning rules or in constraints being derived. 
Therefore, it provides a measure for the utility of the encoded rules in $\mathcal{K}$. 
The action pruning rate describes the ratio of actions that were pruned due to rules firing in the knowledge base.
Therefore, it provides a measure for the direct influence of the encoded rules on the training and application of the KGRL algorithm.

Table \ref{tab:results:exp2} shows $\mathcal{K}$-utilization and $A$-pruning for our KGRL algorithm over all evaluation domains.
In general, the $\mathcal{K}$-utilization is not trivial, indicating that $\mathcal{K}$ contributes frequently to the decision-making of KGRL. 
We could show that, especially for the \textit{HardMoveX} domains, $\mathcal{K}$ caused pruning almost half of the actions, which leads to a large simplification of the decision space. 
While $A$-pruning at \textit{CatchPoint} and \textit{\textit{HardGoal}} ranged around a third, we could show that in the \textit{Platform} domain mainly parameter constraints were deduced from $\mathcal{K}$. 

\begin{table}[h!]
\centering
\caption{The table shows the mean $\mathcal{K}$-utilization $\pm$ one standard deviation and the mean $A$-pruning $\pm$ one standard deviation across eight seeds, for our KGRL algorithm on all evaluation domains.}
\label{tab:results:exp2}
\begin{tabular}{@{}lrll@{}}
\toprule
domain                       & n rules   & $\mathcal{K}$-utilization & $A$-pruning \\ \midrule
HardMove4                    & 48        & 1.00 $\pm$ 0.000 & 0.49 $\pm$ 0.001 \\
HardMove8                    & 96        & 1.00 $\pm$ 0.000 & 0.48 $\pm$ 0.003 \\
HardMove10                   & 120       & 1.00 $\pm$ 0.000 & 0.49 $\pm$ 0.002 \\
CatchPoint                   & 4         & 0.64 $\pm$ 0.078 & 0.32 $\pm$ 0.039 \\
HardGoal                     & 70        & 1.00 $\pm$ 0.000 & 0.34 $\pm$ 0.000 \\
Platform                     & 15        & 0.49 $\pm$ 0.004 & 0.00 $\pm$ 0.000 \\ \bottomrule
\end{tabular}
\end{table}

\begin{figure}[h!]
\centering
\includegraphics[width=0.9\linewidth]{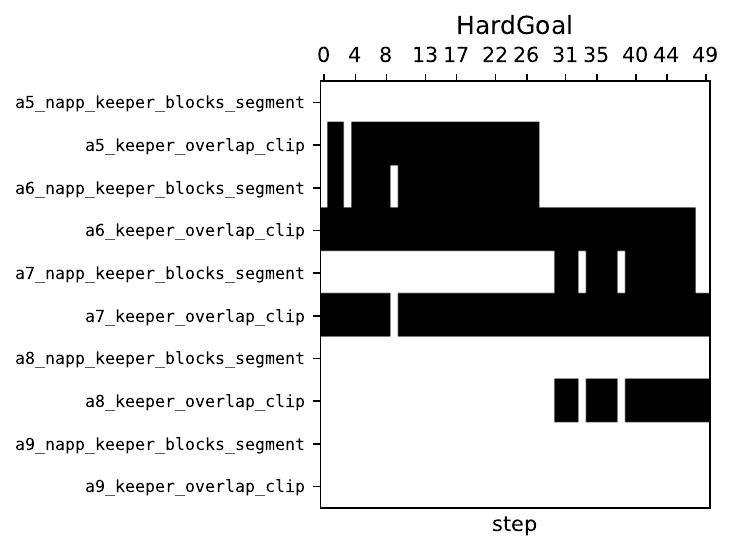}
\caption{The Figure shows an exemplary excerpt from the \textit{HardGoal} domain's explanation trace.
During different phases of the rollout, different $\texttt{napp}(a)$ rules fired due to the goalkeeper blocking them for the agent.} \label{fig:exp2:expl-traces}
\end{figure}

As we can record individual rule activations in $\mathcal{K}$, e.g., $\texttt{napp}(a)$ or $\texttt{active}(a)$, we can construct step-wise explanation traces which explain why certain actions were excluded and how feasible parameter regions were shaped.
This directly supports our RQ3 by showing that the information which is required for rule-grounded explanations is constructed at decision-time. 
Figure \ref{fig:exp2:expl-traces} shows an exemplary excerpt from the explanation trace of the \textit{HardGoal} domain.

\section{Discussion} \label{sec:discussion}

\paragraph{Knowledge Base}
While our evaluation is consistent with the underlying hypothesis of RQ1, that pruning non-applicable actions simplifies the decision space during exploration and deployment, leading to a more sample-efficient training, our ablation study showed that it is important \textit{when} the knowledge base is integrated. 
While removing the knowledge base during evaluation, the performance of KGRL strongly degraded across all domains. 
Removing the knowledge base during training showed a slight negative effect across all domains, except for the \textit{HardGoal} domain.
This shows that the knowledge base mainly affects the deployment of the RL agent, whereas its regularization effect on the agent during training is negligible or even leads to increase in performance.
Through the hard integration in the form of action pruning and hard parameter constraints, our KGRL algorithm additionally is more sensitive to incorrect knowledge and rules.
Wrong rules can harm the performance of the agent by pruning optimal actions or parameter ranges. 
A way of mitigating this risk lies either in using self repairing adaptive mechanisms as suggested in~\cite{Feng2025} or by integrating the knowledge regularization softly in the form of a symbolic policy augmentation as in~\cite{Veronese2026}.

\paragraph{$\textsc{ParamOpt}$}
Our evaluation also underlines the underlying hypothesis of RQ2 that integrating a gradient-guided parameter refinement can increase the sample efficiency and performance of an RL agent compared to one shot estimators. 
This supports our idea that a DQN contains useful gradient information and that a local optimization at decision time outperforms one-shot estimators.
However, analogously to integrating the knowledge-base, it is also important \textit{when} the gradient-guidance is integrated. 
Our ablations clearly showed that integrating the gradient-guidance during training has a negative effect on the performance of our algorithm.
This can be explained by additional noise that is introduced by the gradient-guided refinement during training, when the DQN is not already well-fitted on the underlying PAMDP.
One way of mitigating the risk of negative influence of $\textsc{ParamOpt}$ during training is by using curriculum learning for scheduling its application in training only when sufficient information has been already learnt by the DQN. 
This additionally reduces computational cost when training a KGRL agent. 

As $\textsc{ParamOpt}$ performs projected gradient ascent over a non-convex DQN, there is no guarantee that $\hat{\psi}$ will refine into a global optimum.
A way of mitigating this risk, are strategies like ensemble approaches with differently seeded optimizers, multi-start optimization with different initial guesses, or a combination of both.

\paragraph{Explanation Traces}
Our results show that the explanation traces produced by KGRL are not only an auxiliary logging artifact, but a direct consequence of the decision mechanism of the RL agent. 
As KGRL evaluates the knowledge base and a symbolic state abstraction in every state, it provides an explanation why actions were excluded and why parameters were constrained. 
As the knowledge base utilization is non-trivial across all domains, and leads to substantial pruning in some domains, the reasoning is an integral aspect of KGRL and the recorded explanation traces a non-trivial benefit to its explainability. 
However, in their current form, the explanation traces are limited to local and procedural explanations of individual decisions \cite{Krarup2019}.
An extension to an aggregation of global explanations, e.g. identifying recurring rule activations and decision patterns as post-processing step, is left for future work. 

\paragraph{Learning Knowledge Bases and Abstraction Functions}
Although we worked with handcrafted knowledge bases and abstraction functions, KGRL is not limited to them and can also operate with learned knowledge bases and abstraction functions, as long as they operate with the same symbolic vocabulary.

Approaches for learning the knowledge base can be derived, e.g., from \cite{Law2020,Mazzi2023,Meli2024} which creates answer set programs (that are closely related to Datalog) from examples and background knowledge or by parsing learned preconditions from action model learning \cite{Arora2018} into a knowledge base.
Approaches for learning the abstraction function can comprise algorithms described in~\cite{Ciatto2024,Mao2019,Han2026} or in the most basic variant, any classification algorithm that can map numeric states to symbolic state representations. 

Whether handcrafted or learned, it is worth mentioning that a wrongly learned or crafted abstraction functions carry the risk of incorrectly deriving non-applicable actions and constraints, which can ultimately lead to incorrect pruning of the decision space.
For our threshold-based abstraction functions, this risk increases near the thresholds, especially in noisy environments.

\section{Conclusion} \label{sec:conclusion}
In this paper, we studied Reinforcement Learning in PAMDPs with access to explicit but incomplete domain knowledge. 
We proposed our novel KGRL algorithm, a Neuro-Symbolic extension to DQN that evaluates a Datalog knowledge base and a symbolic state representation at each decision step to prune non-applicable actions and uses projected gradient ascent for a gradient-guided parameter refinement.
Compared to other Neuro-Symbolic RL approaches, KGRL is the first to extend to domains with parametrized action spaces that can be modeled by PAMDPs. 

KGRL outperforms all baseline algorithms on classical PAMDP benchmarks.
We could show that, especially during deployment, the combination of knowledge-guided action pruning and gradient-guided parameter refinement leads to faster convergence on higher episodic returns with lower variance, while additionally providing local, procedural explanation traces via logged rule activations. 
This lets us answer the central question of the paper, whether symbolic domain knowledge can shape the training of RL to be more sample-efficient, positively.
However, we also found that is important \textit{when} to inject \textit{what} knowledge.
Future work will include algorithmic solutions for learning the abstraction function and knowledge base, as well as algorithmic solutions for aggregating and evaluating the explanation traces into global explanations.

\section*{Acknowledgements}
This research as part of the project LaiLa and EKI is funded by dtec.bw – Digitalization and Technology Research Center of the Bundeswehr, which we gratefully acknowledge. 
Computational resources (HPC cluster HSUper) have been provided by the project hpc.bw.
dtec.bw is funded by the European Union – NextGenerationEU.

\bibliographystyle{kr}
\bibliography{references}

\end{document}